\documentclass[journal]{IEEEtran}

\usepackage{amsmath,amsfonts,amssymb,bm}
\usepackage{graphicx}
\usepackage{booktabs}
\usepackage{cite}
\usepackage{xcolor}
\usepackage{orcidlink}
\usepackage{multirow}
\def\x{{\mathbf x}}

\title{Beyond Effective Sample Size: Effective Number of Proposals for Adaptive Importance Sampling}

\author{
	Ali Mousavi$^{\orcidlink{0000-0002-4084-6102}}$ and V\'ictor Elvira$^{\orcidlink{0000-0002-8967-4866}}$, \IEEEmembership{Senior Member, IEEE}
	\thanks{
		A. Mousavi is with the Department of Computer Engineering, Islamic Azad University,
		Neyshabur Branch, Neyshabur, Iran. (e-mail: mousavi@iau.ac.ir).
	}
	\thanks{
		V. Elvira is with the School of Mathematics, University of Edinburgh,
		United Kingdom (e-mail: victor.elvira@ed.ac.uk).
	}
}

\begin{document}
	\markboth{IEEE SIGNAL PROCESSING LETTERS, Vol. XX, No. X, XXXX}
	{Shell \MakeLowercase{\textit{et al.}}: Bare Demo of IEEEtran.cls for IEEE Journals}
	\maketitle
	

	\begin{abstract}
		Population-based adaptive importance sampling (AIS) methods use a set of
		proposal densities to approximate complex target distributions. Their
		performance is commonly assessed through effective sample size (ESS) and related
		weight-based diagnostics, which measure the concentration of normalized
		importance weights. However, a large ESS only indicates that the normalized
		sample weights are not strongly concentrated; it does not describe how the
		proposal components are arranged in the sampling space. In population-based AIS,
		several proposal components may generate samples in the same region of the
		target, so the sample weights can appear well balanced even though the effective
		number of distinct proposal components is small. This letter introduces the
		effective number of proposals (ENP), a similarity-aware proposal-level diagnostic
		for population-based AIS. ENP combines the total normalized weight assigned to
		each proposal with a redundancy measure computed from similarities among
		target-weighted samples, estimating the number of non-redundant empirical
		proposal contributions to the approximation. We establish basic effective-number
		properties and show that ENP detects proposal collapse and duplication missed by
		standard ESS. We also illustrate its use as a targeted feedback signal for
		proposal rejuvenation.
	\end{abstract}

	
	\section{Introduction}
	
In statistical signal processing, many problems require the computation of
expectations with respect to a probability density function. When the target
distribution is complex or known only up to a normalizing constant, Monte Carlo
methods are commonly used to approximate these expectations \cite{liu2001monte}.
Importance sampling (IS) performs this approximation by drawing samples from a
proposal density and correcting the mismatch with importance weights. Since the
efficiency and reliability of IS depend strongly on the proposal distribution,
poor proposal choices can lead to unstable estimators \cite{owen2000safe}.

Multiple importance sampling (MIS) extends IS by combining samples generated
from several proposal distributions, with the balance heuristic being a classical
and widely used weighting strategy \cite{veach1995optimally,owen2000safe,elvira2019gmis}.
Adaptive importance sampling (AIS) methods update the proposal population over
iterations in order to improve the approximation of the target distribution
\cite{bugallo2017adaptive,martino2015adaptive,martino2017layered}. Population Monte Carlo (PMC) is a
representative population-based AIS methodology \cite{cappe2004population,cappe2008adaptive,elvira2017improving}. Related adaptive multiple-proposal methods, such as AMIS,
O-PMC, HAIS, GRAMIS, and HPMC, have explored different mechanisms for adapting
or improving proposal populations \cite{cornuet2012adaptive,elvira2022optimized,
	mousavi2021hamiltonian,elvira2023gradient,mousavi2025hybrid}.
	
A standard diagnostic in the IS family is the effective sample size (ESS).
Conceptually, ESS is interpreted as the number of samples that would need to be
drawn directly from the target distribution to obtain the same variance as the
self-normalized IS estimator under study \cite{kong1992note,liu2001monte}. Since this
variance-based quantity is generally not available in closed form, it is commonly
approximated by empirical diagnostics based on the normalized importance weights.
ESS is also used as an operational signal in
adaptive Monte Carlo algorithms, for instance to decide when resampling should
be performed in sequential Monte Carlo and to improve the robustness of
covariance adaptation in AIS \cite{doucet2012adaptive,el2018robust}.
Several related effective-number diagnostics have also been studied, including
entropy-based perplexity \cite{cover1991elements,cappe2008adaptive}, R\'enyi/Hill-type ESS
families such as the Huggins--Roy family \cite{huggins2015convergence}, and
discrepancy-based ESS measures \cite{martino2017effective}. These measures provide
different ways of quantifying how many effective weighted samples are available.
However, they remain primarily weight-based diagnostics and do not explicitly
account for proposal labels or the similarity structure among samples
\cite{elvira2022rethinking}.

	In MIS and population-based AIS, the use of ESS is more subtle because the
	samples are no longer generated from a single proposal distribution. Instead,
	several proposal densities contribute to the approximation, and different
	sampling, weighting, and adaptation schemes may lead to different statistical
	behavior \cite{elvira2019gmis,bugallo2017adaptive}. In this direction, Elvira et al.
	revisited the classical ESS approximation, highlighted its assumptions and
	limitations, and emphasized that weight-only diagnostics can miss important
	information contained in the integrand, the sample locations, and spatial
	partitions of the sample population \cite{elvira2022rethinking}. Other diagnostic
	developments, such as Pareto-smoothed importance sampling (PSIS), also aim to
	improve the reliability assessment of IS estimators by stabilizing extreme
	importance weights and providing convergence and effective-sample-size
	diagnostics \cite{vehtari2024pareto}. However, these diagnostics still do not
	directly answer a proposal-level question that is central in population-based
	AIS: how many non-redundant proposal components effectively contribute to the
	approximation?
	
	In population-based AIS, the quality of the approximation depends not only on
	the distribution of the importance weights, but also on the configuration of the
	proposal population. Two runs may have comparable ESS values while relying on
	very different proposal arrangements. In particular, several proposal components
	may overlap strongly or collapse to the same region of the target. In such a
	situation, the weighted sample population may still appear reliable according to
	standard weight-based diagnostics, whereas the proposal population has
	effectively lost diversity. Therefore, besides asking how many effective
	weighted samples are available, it is also useful to ask how many
	non-redundant proposal components effectively contribute to the approximation.
	
	Motivated by this observation, this letter introduces the effective
	number of proposals (ENP), a proposal-level diagnostic for AIS. ENP combines
	the total normalized weight of each proposal with a similarity-based redundancy
	measure computed from the weighted samples generated by the proposals, aiming to
	detect proposal collapse or excessive empirical redundancy that may not be
	revealed by classical weight-based ESS. The main contributions are:
	\begin{itemize}
		\item We define a similarity-aware ENP for measuring the effective number of
		non-redundant proposal components.
		\item We establish its basic effective-number properties under
		non-redundant, collapsed, and duplicated proposal configurations.
		\item We show that ENP detects proposal redundancy missed by ESS and provides
		a useful signal for proposal rejuvenation.
	\end{itemize}
	The remainder of this letter is organized as follows. Section~II gives the
	background and notation. Section~III presents the proposed ENP and its basic
	properties. Section~IV reports the numerical experiments, and Section~V
	concludes the letter.
	\section{Background}
	This section introduces the notation for deterministic-mixture importance
	sampling and the weight-based ESS diagnostics used for comparison.
\subsection{Deterministic-Mixture Importance Sampling}

Let $\pi(\x)$ denote an unnormalized target density on $\mathbb{R}^d$. At a
given iteration of a population-based adaptive importance sampling method, let
$q_1(\x),\ldots,q_N(\x)$ be the current proposal densities. Each proposal
$q_n(\x)$ generates samples $\x_{n,k}\sim q_n(\x)$, $k=1,\ldots,K$, giving a
total sample size $S=NK$.

In deterministic-mixture importance sampling, also known as balance-heuristic
weighting in the MIS literature \cite{owen2000safe,elvira2019gmis}, the mixture
denominator is

\begin{equation}
	\psi(\x)=\frac{1}{N}\sum_{j=1}^{N}q_j(\x),
\end{equation}
and the unnormalized importance weight of sample $\x_{n,k}$ is
\begin{equation}
	\omega_{n,k}
	=
	\frac{\pi(\x_{n,k})}{\psi(\x_{n,k})}.
\end{equation}
The normalized weights are
\begin{equation}
	w_{n,k}
	=
	\frac{\omega_{n,k}}
	{\sum_{j=1}^{N}\sum_{\ell=1}^{K}\omega_{j,\ell}},
	\qquad
	\sum_{n=1}^{N}\sum_{k=1}^{K}w_{n,k}=1.
\end{equation}
We also use the stacked notation $\bm w=(w_1,\ldots,w_S)$, where each index
$i$ corresponds to a pair $(n,k)$.

	\subsection{Weight-Based ESS Diagnostics}
	The classical empirical approximation of ESS is based on the normalized
	importance weights and is defined as
	\begin{equation}
		\widehat{\mathrm{ESS}}_{2}(\bm w)
		=
		\frac{1}{\sum_{i=1}^{S}w_i^2}.
		\label{eq:ess2}
	\end{equation}
	This diagnostic is large when the normalized weights are close to uniform and
	small when a few samples dominate \cite{kong1992note,liu2001monte,elvira2022rethinking}. It is
	therefore useful for detecting weight degeneracy. However, it does not use
	sample locations, proposal labels, or similarities among samples.
    Several alternative ESS-like metrics have been proposed in the literature. A
	well-known example is the perplexity, which is based on the discrete entropy of
	the normalized weights \cite{cover1991elements,cappe2008adaptive},
	\begin{equation}
		\mathrm{Per}(\bm w)
		=
		\exp
		\left(
		-\sum_{i=1}^{S}w_i\log w_i
		\right).
		\label{eq:perplexity}
	\end{equation}
	Another useful alternative is the inverse-maximum-weight ESS \cite{martino2017effective,elvira2022rethinking},
	\begin{equation}
		\widehat{\mathrm{ESS}}_{\infty}(\bm w)
		=
		\frac{1}{\max_i w_i}.
		\label{eq:ess_infty}
	\end{equation}
	This measure is more conservative than \eqref{eq:ess2},  it is controlled
	by the most dominant sample weight. 
	More generally, ESS-like diagnostics can be interpreted through
	R\'enyi/Hill-type effective-number families \cite{huggins2015convergence,
		elvira2022rethinking}. In this view, the classical ESS, perplexity, and
	inverse-maximum-weight ESS correspond to different orders of the same
	weight-based effective-number principle.
	
	These metrics provide different effective-number interpretations of the
	normalized weights. Nevertheless, they remain functions of the weight vector
	alone. Therefore, they do not explicitly account for the geometric locations of
	the samples, the proposal labels, or the overlap between proposal components.

	\section{Effective Number of Proposals}
	\label{sec:enp}
	
	We now define the proposed proposal-level diagnostic. While ESS quantifies the
	effective number of weighted samples, the goal here is to quantify how many
	non-redundant proposal components effectively contribute to the AIS
	approximation. The construction has two ingredients: a similarity-aware proposal
	redundancy measure and a redundancy-corrected effective number. Since ENP
	identifies when the proposal population becomes redundant, it can also be used
	as a feedback signal for adaptation.
\subsection{Similarity-Aware Proposal Redundancy}
\label{subsec:proposal_redundancy}

The key idea of ENP is to compare, for each sample, the weighted mass around it
in the whole sample population with the weighted mass around it generated by its
own proposal. If these two quantities are close, the proposal is locally
distinct; if the whole population contains much more nearby mass, the proposal
has redundant target-weighted contributions with other proposals. Thus, ENP
measures empirical, target-weighted redundancy rather than analytical distances
between proposal densities.

Let $Z\in[0,1]^{S\times S}$ be a symmetric sample-similarity matrix with
$Z_{ii}=1$, motivated by similarity-sensitive diversity measures
\cite{leinster2012measuring}. The entry $Z_{ij}$ measures the similarity between samples
$\x_i$ and $\x_j$. For a sample $\x_i$, define its similarity-weighted
neighborhood mass as
\begin{equation}
	(Z\bm w)_i
	=
	\sum_{j=1}^{S} Z_{ij}w_j .
	\label{eq:ordinariness}
\end{equation}
This quantity is large when $\x_i$ lies near many high-weight samples and small
when it is relatively isolated.

The total normalized weight assigned to proposal $n$ is
\begin{equation}
	v_n
	=
	\sum_{k=1}^{K} w_{n,k},
	\qquad
	\sum_{n=1}^{N}v_n=1.
\end{equation}
For $v_n>0$, define the locally normalized weights
\begin{equation}
	\tilde w_{n,k}
	=
	\frac{w_{n,k}}{v_n},
	\qquad
	k=1,\ldots,K.
\end{equation}

Let $Z^{(n)}$ denote the $K\times K$ block of $Z$ associated with the samples
generated by proposal $n$. For sample $\x_{n,k}$, the global neighborhood mass is
$(Z\bm w)_{n,k}$, while the neighborhood mass due only to proposal $n$ is
$v_n(Z^{(n)}\tilde{\bm w}_n)_k$. We define the sample-wise redundancy ratio as
\begin{equation}
	R_{n,k}
	=
	\frac{
		(Z\bm w)_{n,k}
	}
	{
		v_n (Z^{(n)}\tilde{\bm w}_n)_k
	}.
	\label{eq:ratio}
\end{equation}
Thus, $R_{n,k}\approx 1$ means that the neighborhood of $\x_{n,k}$ is mainly
explained by proposal $n$ itself, whereas $R_{n,k}>1$ indicates overlap with
samples generated by other proposals.

For a general order $q$, we define the proposal redundancy as a weighted power
mean of the ratios:
\begin{equation}
	\rho_n^{(q)}
	=
	\left(
	\sum_{k=1}^{K}
	\tilde w_{n,k}
	R_{n,k}^{q-1}
	\right)^{\frac{1}{q-1}},
	\qquad q\neq 1.
	\label{eq:rho_general}
\end{equation}
For $q=1$, the continuous limit is the weighted geometric mean:
\begin{equation}
	\rho_n^{(1)}
	=
	\prod_{k=1}^{K}
	R_{n,k}^{\tilde w_{n,k}}
	=
	\exp
	\left(
	\sum_{k=1}^{K}
	\tilde w_{n,k}\log R_{n,k}
	\right).
	\label{eq:rho_one}
\end{equation}
For $q=2$, the redundancy reduces to the weighted arithmetic mean:
\begin{equation}
	\rho_n^{(2)}
	=
	\sum_{k=1}^{K}
	\tilde w_{n,k}R_{n,k}.
\end{equation}

Thus, $\rho_n^{(q)}$ is a proposal-level redundancy index. Values close to one
indicate that proposal $n$ contributes samples that are mostly distinct from the
rest of the population. Larger values indicate stronger overlap with other
proposals.
	
	\subsection{Redundancy-Corrected ENP}
	
	First, define the bounded redundancy
	\begin{equation}
		\bar{\rho}_n^{(q)}
		=
		\min
		\left\{
		\rho_n^{(q)},\frac{1}{v_n}
		\right\},
		\qquad v_n>0.
		\label{eq:bounded_rho}
	\end{equation}
	Thus,
	\begin{equation}
		1
		\leq
		\bar{\rho}_n^{(q)}
		\leq
		\frac{1}{v_n}.
	\end{equation}
	This upper bound prevents a highly redundant proposal from dominating the
	diagnostic and ensures that the resulting effective number remains within the
	natural range $[1,N]$.
	
	Inspired by Hill/R\'enyi effective numbers and generalized ESS families, we
	define the following family of redundancy-corrected effective numbers of
	proposals:
	\begin{equation}
		\mathrm{ENP}_{\rho}^{(r,q)}
		=
		\left(
		\sum_{n:v_n>0}
		v_n^{r}
		\left(\bar{\rho}_n^{(q)}\right)^{r-1}
		\right)^{\frac{1}{1-r}},
		\qquad r>1.
		\label{eq:enp_family}
	\end{equation}
	Here, $r$ is the effective-number order and $q$ is the redundancy-diversity
	order used in $\rho_n^{(q)}$. Larger values of $r$ give more emphasis to
	dominant proposal weights, while the redundancy term
	$\bar{\rho}_n^{(q)}$ penalizes proposals that overlap with the rest of the
	population.
	
	In this letter, we focus on the order-$2$ member of the family, which gives the
	collision-type ENP
	\begin{equation}
		\mathrm{ENP}_{\rho}^{(2,q)}
		=
		\left(
		\sum_{n:v_n>0}
		v_n^2 \bar{\rho}_n^{(q)}
		\right)^{-1}.
		\label{eq:enp}
	\end{equation}
	The denominator in \eqref{eq:enp} can be interpreted as a redundancy-corrected
	proposal collision probability. The term $v_n^2$ is the probability that two
	independent weighted draws select proposal $n$, while
	$\bar{\rho}_n^{(q)}$ inflates this collision contribution when proposal $n$ is
	redundant with the rest of the population.
	
	\subsection{Basic Properties}
	\label{subsec:enp_properties}
	
	The proposed diagnostic satisfies natural effective-number properties. Since
	\begin{equation}
		1
		\leq
		\bar{\rho}_n^{(q)}
		\leq
		\frac{1}{v_n},
		\qquad v_n>0,
	\end{equation}
	we have
	\begin{equation}
		\sum_{n:v_n>0} v_n^2
		\leq
		\sum_{n:v_n>0} v_n^2 \bar{\rho}_n^{(q)}
		\leq
		\sum_{n:v_n>0} v_n
		=
		1.
	\end{equation}
	Taking reciprocals gives
	\begin{equation}
		1
		\leq
		\mathrm{ENP}_{\rho}^{(2,q)}
		\leq
		\frac{1}{\sum_{n:v_n>0}v_n^2}
		\leq
		N.
		\label{eq:enp_range}
	\end{equation}
	Thus, ENP lies between one and the nominal number of proposals.
	
	The limiting cases are also consistent with an effective-number interpretation.
	If the proposals are non-redundant, then $\rho_n^{(q)}\approx 1$ and
	\begin{equation}
		\mathrm{ENP}_{\rho}^{(2,q)}
		\approx
		\frac{1}{\sum_{n=1}^{N}v_n^2},
	\end{equation}
	which is the $\widehat{\mathrm{ESS}}$ of the proposal-weight vector
	$\bm v=(v_1,\ldots,v_N)$. In particular, if the non-redundant proposals are
	uniformly weighted, $v_n=1/N$, then
	$\mathrm{ENP}_{\rho}^{(2,q)}\approx N$. Conversely, if a single proposal
	carries all the weight, then $\mathrm{ENP}_{\rho}^{(2,q)}=1$.
	
	The diagnostic is also sensitive to proposal redundancy. If all $N$ proposals
	have equal weights and generate highly overlapping samples, then
	$v_n=1/N$ and $\rho_n^{(q)}\approx N$, yielding
	\begin{equation}
		\mathrm{ENP}_{\rho}^{(2,q)}
		\approx
		\left(
		N
		\left(\frac{1}{N}\right)^2
		N
		\right)^{-1}
		=
		1.
	\end{equation}
	More generally, suppose that $N=ML$ proposals form $M$ distinct proposal
	regions, each duplicated $L$ times, with uniform weights. Then
	$v_n=1/(ML)$ and $\rho_n^{(q)}\approx L$, so
	\begin{equation}
		\mathrm{ENP}_{\rho}^{(2,q)}
		\approx
		\left(
		ML
		\left(\frac{1}{ML}\right)^2
		L
		\right)^{-1}
		=
		M.
	\end{equation}
	Therefore, duplicating already existing proposal components does not
	artificially increase the effective number of proposals.
	
	These properties show that $\mathrm{ENP}_{\rho}^{(2,q)}$ behaves as an
	effective number of non-redundant proposals. The general family in
	\eqref{eq:enp_family} provides additional flexibility, while the order-$2$
	definition in \eqref{eq:enp} is the simplest and most directly connected to the
	classical $\mathrm{ESS}$.
	
	 
	\section{Numerical Experiments}
	\label{sec:experiments}
	
	We consider the standard two-dimensional five-modal Gaussian mixture target used
	in AIS studies. Unless otherwise stated, all proposals are Gaussian with
	isotropic covariance $\sigma_q^2 I_2$. Sample similarities are computed with a
	Gaussian kernel
	$Z_{ij}=\exp(-\|\x_i-\x_j\|^2/(2\sigma_Z^2))$. In all experiments we use $q=2$
	and report $\mathrm{ENP}_{\rho}^{(2,2)}$.
	\subsection{Controlled Diagnostic Experiment}
	\label{subsec:controlled_experiment}
	
	We first evaluate ENP in controlled proposal configurations, isolating the
	diagnostic behavior from the adaptation mechanism. In each scenario, $K=500$
	samples are generated from each proposal and results are averaged over $30$
	independent runs. We consider: (A) five non-redundant proposals, each centered at
	a different target mode; (B) five collapsed proposals, all centered at the same
	target mode; and (C) six proposals forming two distinct groups, each duplicated
	three times. To assess sensitivity to the similarity bandwidth, ENP is computed
	for $\sigma_Z\in\{1.5,2.5,4.0\}$, corresponding to local, moderate, and broad
	similarity scales relative to the proposal standard deviation $\sigma_q=1.5$.
	Table~\ref{tab:controlled_scenarios} shows that weight-based diagnostics may
	remain large even when the proposal population is redundant. In Scenario A, ENP
	remains close to five for all tested bandwidths. In Scenario B, the ESS and
	perplexity rates remain relatively high, whereas ENP correctly decreases to
	approximately one. In Scenario C, ENP is approximately two, matching the number
	of distinct proposal groups. These results show that ENP is stable for bandwidths
	chosen around the proposal scale and counts effective non-redundant proposal
	components rather than proposal labels.
	
\subsection{Diagnostic-Triggered Proposal Rejuvenation}
\label{subsec:diagnostic_rejuvenation}

We finally illustrate that ENP can be used as an adaptation signal from the
collapsed initialization on the same five-modal target. We use $N=50$ Gaussian
proposals, $K=20$ samples per proposal, and $T=40$ adaptive DM-PMC iterations,
with $\sigma_q=2$ and $\sigma_Z=2.5$. Results are averaged over $20$
independent runs. For the adaptive experiment we use $\sigma_Z=2.5$.
We compare standard DM-PMC with two diagnostic-triggered rejuvenation variants.
Motivated by the common use of ESS-based thresholds as resampling or adaptation
criteria in sequential and adaptive Monte Carlo methods
\cite{liu1998sequential,bugallo2017adaptive,elvira2017improving}, we compare ESS-triggered and
ENP-triggered rejuvenation rules. 
In the ESS-triggered variant, proposal rejuvenation is applied when
$\widehat{\mathrm{ESS}}_2/S<0.5$, where $S=NK$. In the ENP-triggered variant,
rejuvenation is applied when $\mathrm{ENP}_{\rho}^{(2,2)}<2.5$. In both
variants, $40\%$ of the proposal locations are rejuvenated. New locations are
drawn either uniformly over the sampling domain $[-20,20]^2$ or from high-weight
samples with inflated Gaussian perturbations. For the ENP-triggered variant, the
rejuvenated proposals are selected as those with the largest
redundancy-corrected contributions $v_n^2\bar\rho_n^{(2)}$.

\begin{table}[t]
	\centering
	\caption{Controlled diagnostic scenarios on the two-dimensional five-modal
		target. The rates ESS$_2/S$, Per./$S$, and ESS$_\infty/S$ are normalized by the
		total number of samples $S=NK$. ENP is reported for different similarity
		bandwidths $\sigma_Z$.}
	\label{tab:controlled_scenarios}
	\begin{tabular}{lcccccc}
		\toprule
		\multirow{2}{*}{Sc.}
		& \multirow{2}{*}{ESS$_2/S$}
		& \multirow{2}{*}{Per./$S$}
		& \multirow{2}{*}{ESS$_\infty/S$}
		& \multicolumn{3}{c}{ENP$_{\rho}^{(2,2)}$} \\
		\cmidrule(lr){5-7}
		& & &
		& $\sigma_Z=1.5$
		& $\sigma_Z=2.5$
		& $\sigma_Z=4.0$ \\
		\midrule
		A
		& 0.794
		& 0.860
		& 0.196
		& 4.998
		& 5.000
		& 4.961 \\
		B
		& 0.733
		& 0.794
		& 0.536
		& 1.005
		& 1.000
		& 1.002 \\
		C
		& 0.790
		& 0.850
		& 0.280
		& 2.005
		& 2.000
		& 2.001 \\
		\bottomrule
	\end{tabular}
\end{table}

\begin{table}[t]
	\centering
	\caption{Diagnostic-triggered proposal rejuvenation from collapsed
		initialization. Coverage is the number of reached target modes; MSE is computed
		for the self-normalized target-mean estimate.}
	\label{tab:diagnostic_rejuvenation}
	\begin{tabular}{lccccc}
		\toprule
		Method & ESS$_2/S$ & ENP & Coverage & MSE & Trig. \\
		\midrule
		Standard DM-PMC
		& 0.618
		& 4.47
		& 3.60
		& 46.48
		& 0.00 \\
		ESS-triggered
		& 0.459
		& 5.93
		& 4.75
		& 13.46
		& 22.00 \\
		ENP-triggered
		& 0.603
		& 6.03
		& 4.90
		& 2.87
		& 1.15 \\
		\bottomrule
	\end{tabular}
\end{table}

Table~\ref{tab:diagnostic_rejuvenation} shows that ENP provides a more targeted
adaptation signal. ESS-triggered rejuvenation improves mode coverage but
requires frequent intervention. In contrast, ENP-triggered rejuvenation achieves
nearly full coverage, the lowest MSE, and only $1.15$ triggers on average. This
suggests that ENP responds directly to proposal redundancy and can guide
proposal rejuvenation more efficiently than a purely weight-based trigger.

	\section{Conclusion}
	
	This letter introduced the effective number of proposals (ENP), a
	similarity-aware diagnostic for measuring proposal-level redundancy in
	population-based adaptive importance sampling. Unlike ESS, which measures the
	effective number of weighted samples, ENP estimates the number of
	non-redundant proposal components contributing to the approximation. The
	proposed diagnostic satisfies natural effective-number properties and behaves
	consistently under non-redundant, collapsed, duplicated, and weight-degenerate
	proposal configurations. Numerical experiments showed that ENP can reveal
	proposal collapse even when standard weight-based diagnostics remain large.
	Moreover, an ENP-triggered rejuvenation experiment showed that ENP can be used
	as a targeted feedback signal for improving proposal adaptation.
	
\bibliographystyle{IEEEtran}
\bibliography{enp_references}
	
\end{document}